\documentclass[journal]{IEEEtran}
\ifCLASSINFOpdf
   \usepackage[pdftex]{graphicx}
\else
\fi
\ifCLASSOPTIONcompsoc
  \usepackage[caption=false,font=normalsize,labelfont=sf,textfont=sf]{subfig}
\else
  \usepackage[caption=false,font=footnotesize]{subfig}
\fi
\usepackage{multirow}
\usepackage{tabularx}

\begin{document}
%
\title{Deep Multimodality Learning for \\
UAV Video Aesthetic Quality Assessment}
%
%
%

\author{Qi~Kuang,
        Xin~Jin,
        Qinping~Zhao,
        Bin~Zhou
\thanks{Manuscript received April 11, 2019; revised August 2, 2019 and October 28, 2019; accepted December 5, 2019. This work was supported in part by National Natural Science Foundation of China (U1736217 and 61932003), National Key R\&D Program of China (2019YFF0302902), Pre-research Project of the Manned Space Flight (060601),  the Open Project Program of State Key Laboratory of Virtual Reality Technology and Systems, Beihang University (VRLAB2019C03), and National Defense Technology Innovation Program of China.}
\thanks{Q. Kuang and Q. Zhao are with State Key Laboratory of Virtual Reality Technology and Systems, School of Computer Science and Engineering, Beihang University, Beijing, China e-mail: (kuangqi@buaa.edu.cn; zhaoqp@buaa.edu.cn).}
\thanks{X. Jin is with Beijing Electronic Science and Technology Institute, Beijing, China e-mail: (jinxin@besti.edu.cn).}
\thanks{B. Zhou is with State Key Laboratory of Virtual Reality Technology and Systems, School of Computer Science and Engineering, Beihang University,Beijing, China, and also with Peng Cheng Laboratory, Shenzhen, China. B. Zhou is the corresponding author. email: (zhoubin@buaa.edu.cn).}
}
%
%


\markboth{Journal of \LaTeX\ Class Files,~Vol.~14, No.~8, August~2015}%
{Kuang \MakeLowercase{\textit{et al.}}: Deep Multimodality Learning for UAV Video Aesthetic Quality Assessment}

%



\maketitle

\begin{abstract}
Despite the growing number of unmanned aerial vehicles (UAVs) and aerial videos, there is a paucity of studies focusing on the aesthetics of aerial videos that can provide valuable information for improving the aesthetic quality of aerial photography. In this article, we present a method of deep multimodality learning for UAV video aesthetic quality assessment. More specifically, a multistream framework is designed to exploit aesthetic attributes from multiple modalities, including spatial appearance, drone camera motion, and scene structure. A novel specially designed motion stream network is proposed for this new multistream framework. We construct a dataset with 6,000 UAV video shots captured by drone cameras. Our model can judge whether a UAV video was shot by professional photographers or amateurs together with the scene type classification. The experimental results reveal that our method outperforms the video classification methods and traditional SVM-based methods for video aesthetics. In addition, we present three application examples of UAV video grading, professional segment detection and aesthetic-based UAV path planning using the proposed method.
\end{abstract}

\begin{IEEEkeywords}
Aesthetic quality assessment, aerial video aesthetic, deep multimodality learning.
\end{IEEEkeywords}

%
\IEEEpeerreviewmaketitle

\section{Introduction}
%
%
%
%
\IEEEPARstart{U}{nmanned} aerial vehicles (UAVs) are used in different areas. One of the most popular applications is photography, which makes it possible for individuals to view the world from novel views in the sky. Aerial photography was a difficult task several years ago and usually required helicopters and professional photographers. With the development of drones (and especially commercial drones), aerial photography does not necessitate expensive equipment. However, the lack of photography knowledge makes it difficult to obtain good quality videos. Thus, it is observed that individuals increasingly focus on UAV video aesthetics, which is significantly related to UAV video quality.

Photo and video aesthetic quality assessment has been popular in recent years~\cite{tian2015query,sheng2018gourmet,zhang2019gated}. Various methods and datasets are designed to exploit video aesthetic features to evaluate video quality~\cite{zhang2018detecting}.
Nevertheless, there is a paucity of studies on UAV video aesthetics, although UAV videos are ubiquitous on the Internet.

In this study, we address the aesthetic quality assessment of UAV videos. Compared with problems of aesthetic quality assessment of images and ordinary videos, there are several differences:

\begin{itemize}

\item To the best of our knowledge, there is no such dataset that contains UAV videos only and is specifically designed for aesthetic quality assessment of UAV videos.

\item Cameras on UAVs always move in the sky. For ordinary videos, cameras are often still.

\item UAV videos are often used for landscape photography. Thus, the scene structures also make contributions to the aesthetics of UAV videos.

\end{itemize}

First, we construct a dataset with 6,000 UAV video shots captured by drone cameras. In the dataset, 3,000 of the video shots are labeled as professional, and the other 3,000 are labeled as amateur. Then, we propose a multistream network that consists of spatial, motion, and structural streams to exploit the multimodal features of photoaesthetics, camera motion, and shooting scene structure.

We modify a pretrained model to extract the features of video frames and employ long-term temporal modeling (LSTM) to take advantage of temporal clues. Subsequently, we use translation and rotation to represent the trajectory of UAV and mounted camera motion, respectively. We propose a network that exploits the characteristics of the UAV trajectory and camera motion and take advantage of the relationship between track points for UAV video aesthetic quality assessment. We also consider the relationship between the structure of scenes and video aesthetics.

For different types of scenes, different photography methods are used. Thus, we optimize two branches with different tasks, namely, aesthetic assessment and scene type classification in the bifurcated subnetwork. The results of our experiments reveal that our method can effectively distinguish professional and amateur videography together with the scene type classification. Our method outperforms video classification methods and traditional SVM-based methods for video aesthetics. We also present three applications of our method, namely, UAV video grading, professional segment detection and aesthetic-based UAV path planning.

The main contributions of this work can be summarized as follows:
\begin{itemize}
\item We construct a dataset containing 6,000 UAV video shots. To the best of our knowledge, this is the first dataset for UAV video aesthetics.
\item A multistream framework consisting of spatial, motion, and structural streams is proposed to exploit comprehensive aesthetic attributes from multiple modalities, including spatial appearance (photo aesthetics), drone camera motion and scene structure.
\item For the motion stream, we propose a novel network that maximizes the relationship between neighboring track points to explore the characteristics of 3D trajectories.
\end{itemize}
\section{Related Works}
We divide the discussions of related works into the following three subsections.
\subsection{Video Classification}
In contrast to deep learning, support vector machines (SVMs) are the dominant classifier option for video classification for over a decade~\cite{lin2002news,suresh2004content,zhou2005svm,yuan2006automatic,brezeale2008automatic}. Recently, neural networks have also been adopted for video classification given the increasing popularity of deep learning-based approaches.

To ensure that the networks perform well, there is a trend wherein the structure of networks is increasingly complex such that more information is exploited~\cite{zhang2019exploiting}. For example, the aim of multiresolution CNN architecture, including multiresolution streams, involves classifying large-scale videos that can correspond to the million-level~\cite{karpathy2014large}. With the exception of raw frame streams, an optical flow stream is introduced to encode the pattern of the apparent motion of objects in a visual scene, and this performs well in classifying action videos~\cite{yue2015beyond}. ~\cite{lan2017deep} trained temporal segment networks using local video snippets as local feature extractors and then aggregated local features to form global features for action recognition. ARTNets are constructed by stacking multiple generic building blocks to simultaneously model appearance and relation from RGB input in a separate and explicit manner for video classification.~\cite{wang2018appearance}
A multilayer and multimodal approach is proposed to capture diverse static and dynamic cues from four highly complementary modalities at multiple temporal scales to incorporate various levels of semantics in every single network~\cite{yang2016multilayer}. Audio features are also augmented by multistream regularized deep neural networks to exploit feature and class relationships~\cite{jiang2018exploiting}.


\subsection{Photoaesthetic Quality Assessment}
Several studies examined image quality based on photoaesthetics~\cite{lu2015rating}. The earliest study on photoaesthetics was published in 2004 to the best of our knowledge.~\cite{tong2004classification} proposed a regression method that can distinguish between photos taken by professional and amateur photographers. Additionally, most subsequent studies involve fitting the results evaluated by humans based on multiple defined handcrafted aesthetic features~\cite{zhang2014fusion}, such as global features~\cite{tong2004classification,ke2006design} and local features~\cite{datta2006studying}. Then, researchers proposed methods based on more aesthetic features, including color harmony~\cite{nishiyama2011aesthetic}, describable attribute characteristics~\cite{dhar2011high} and generic image descriptors~\cite{marchesotti2011assessing}.

Currently, most studies on photoaesthetic quality assessment automatically extract the aesthetic features with deep learning~\cite{lee2017photo}.
~\cite{lu2014rapid} incorporated a global view and a local view of the image and unified the feature learning and classifier training using a double column deep convolutional neural network.
~\cite{kao2015visual} utilized a convolutional network to extract the aesthetic features that are difficult to design manually, and then a regression model was trained based on the aesthetic features, which can predict a continuous aesthetic score.
~\cite{lu2015deep} trained a deep multipath aggregation network using multiple patches generated from one image to solve three problems: image style recognition, aesthetic quality categorization, and image quality estimation.
~\cite{kong2016photo} proposed learning a deep convolutional neural network that incorporates joint learning of meaningful photographic attributes and image content information to rank photoaesthetics.
Various convolutional neural networks for image recognition are modified to assess aesthetics~\cite{ma2017lamp}. Image style, image content and some additional information are modeled explicitly or implicitly with convolutional neural networks. The convolutional neural networks perform better than those methods using handcrafted aesthetic features.~\cite{jin2018ilgnet} proposed ILGNet, which exhibits excellent performance on the AVA dataset, which is a large-scale database for aesthetic visual analysis~\cite{murray2012ava}.

\subsection{Video Aesthetic Quality Assessment}
In contrast to image aesthetic quality assessment~\cite{JinAAAI2018}, only a few video aesthetic quality assessment methods have been proposed to date. A frequently used method initially defines a few features associated with the aesthetics of video. Photoaesthetics include several rules collected from professionals for amateurs to follow, and thus, the criterion of photoaesthetics is introduced to videos and include color saturation, focus control~\cite{luo2008photo}, and exposure~\cite{niu2012makes}. Additionally, the objective involves assessing the aesthetic quality of videos, and thus, video features including visual continuity, camera motion, and shooting type~\cite{yeh2013video} are used as the inputs of SVM with image aesthetic features.

Given that there is a paucity of studies on video aesthetic quality assessment using deep learning, existing datasets for video aesthetic quality assessment are not as abundant as those of images. The authors in~\cite{niu2012makes} made a dataset that included 1,000 professional shots collected from 16 feature movies and 34 commercial TV shows, and 1,000 amateur shots taken by 23 amateur users to extract aesthetic features.~\cite{yeh2013video} set up an ADCCV dataset that enhanced the Telefonica dataset~\cite{moorthy2010towards} by augmenting it with more positive examples.~\cite{tzelepis2016video} assessed video aesthetic quality via a kernel SVM extension with the CERTH-ITI-VAQ700 dataset, including 350 high aesthetic quality videos and 350 low aesthetic quality videos. To summarize, neural networks are not extensively adopted for video aesthetic quality assessment subjected to small-scale datasets. To the best of our knowledge, there is a paucity of datasets related to UAV video aesthetics, and thus, the present study focuses on determining UAV video aesthetics through deep learning.
\section{Methods}
\begin{figure*}
\begin{center}
  \includegraphics[width=\linewidth]{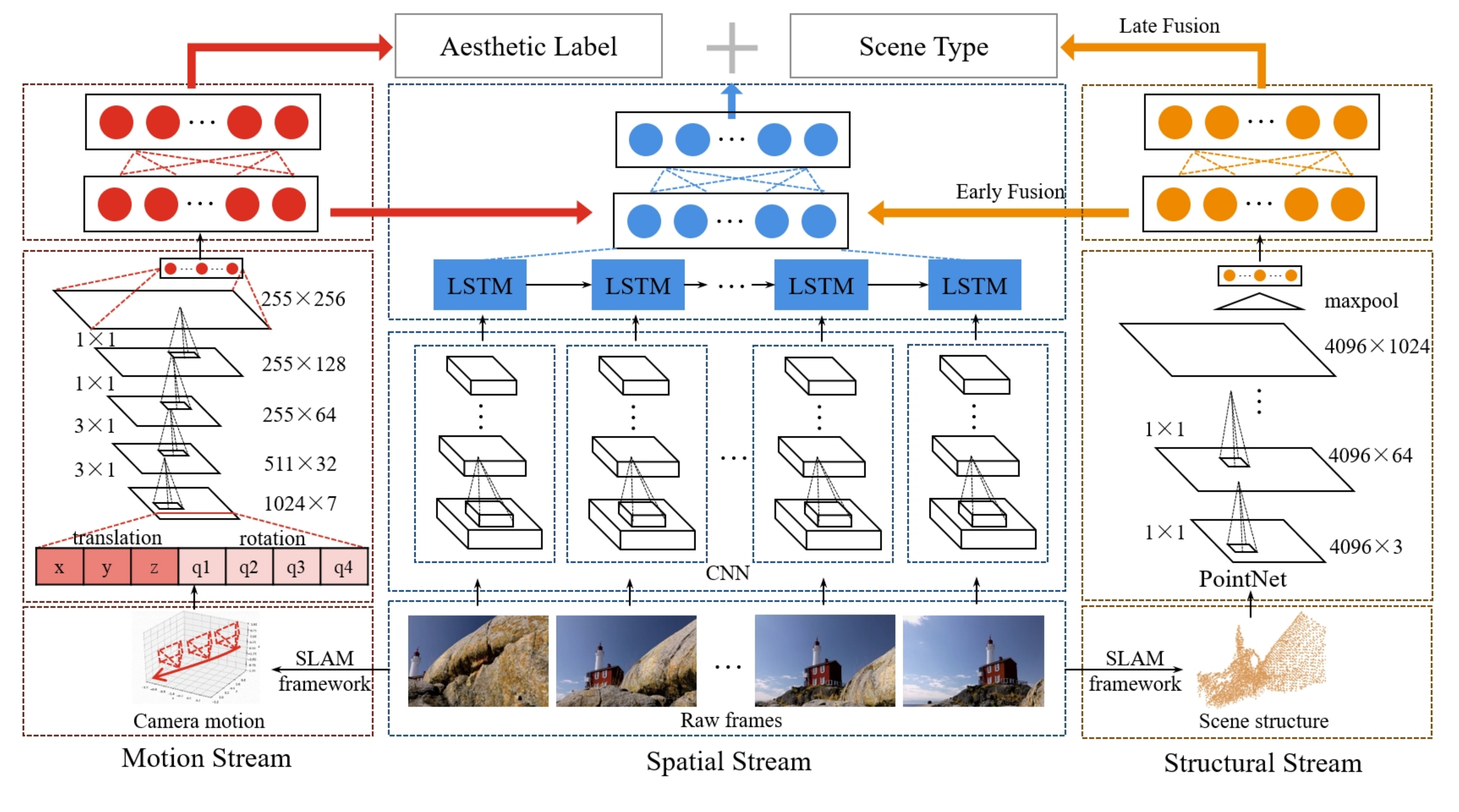}\\
\end{center}
  \caption{Illustration of the proposed framework. We use three streams for exploiting multimodal features. Additionally, we leverage a multitask output, including aesthetic assessment and scene type classification. For different scene types, different photography methods should be used.}
\label{fig1}
\end{figure*}
In this section, we first describe the proposed framework for UAV video aesthetic quality assessment, as shown in Figure~\ref{fig1}, and then introduce individual network streams.

\subsection{Overview of the Framework}
Videos are inherently multimodal, and thus, we introduce a multistream network to gather abundant multimodal information that distinguishes between professional UAV videos and amateur videos based on human thinking. The most intuitive rule to determine UAV video aesthetics corresponds to frame aesthetics. Additionally, we consider camera motion as a significant factor based on the flexibility of drones. Furthermore, the constructed structure of the scene that is shot also reflects the difference in shooting content and shooting type. Based on the recently proposed multistream approach~\cite{wu2016multi,jiang2018exploiting}, we train three convolutional neural networks (ConvNets), namely, spatial, motion, and structural streams, to decompose UAV videos for aesthetic quality assessment.

\subsection{Spatial Stream}
Evidently, video aesthetic quality is significantly associated with photoaesthetic quality. It is unlikely that a video with unattractive frames appears professional, and thus, photoaesthetic features are introduced to the network.~\cite{mai2016composition} proved that a conventional deep convolutional neural network can be applied to photoaesthetic assessment, and the results are promising and impressive. Thus, we reuse a classification ConvNet architecture pretrained on a large collection of images, such as ImageNet~\cite{deng2009imagenet}, such that we can obtain high-dimensional features that can be then used for photoaesthetic assessment. Subsequently, the top layer of the network is modified for our task.

Our task involves video aesthetic quality assessment, and thus, we employ LSTM to model long-term temporal clues. The LSTM can exploit temporal information of a data sequence with an arbitrary length by recursively mapping the input sequence to output labels with hidden units. Therefore, we use LSTM instead of the fully connected layer used as the top layer. With regard to our problem, the input should be the feature of an input video frame.

It should be noted that ConvNet with LSTM was applied for video classification in previous studies. The task of video aesthetic quality assessment involves evaluating whether a video is professional (which is similar to video classification to a certain extent); thus, we define it as our baseline that represents most methods for video classification.

\subsection{Motion Stream and Structural Stream}
Several previous studies on video classification introduced the motion feature to obtain better results. Most of their tasks involve recognizing actions, and thus, optical flows are used to encode subtle motion patterns of an object. With respect to video aesthetic quality assessment, we focus on camera motion. A professional video features good camera motion and especially a UAV video. Based on our experience of taking UAV videos, camera motion is an extremely important element because UAVs are more flexible. Camera motion can be decomposed into translation and rotation under normal conditions. However, a camera mounted on a UAV cannot translate by itself. Thus, camera translational motion actually means the movement of the UAV. We use a three-dimensional vector to represent the trajectory of the UAV and a quaternion to represent the rotation of the camera mounted on the UAV.

\begin{figure}[!t]
\centering
\subfloat[]{\includegraphics[width=0.45\linewidth]{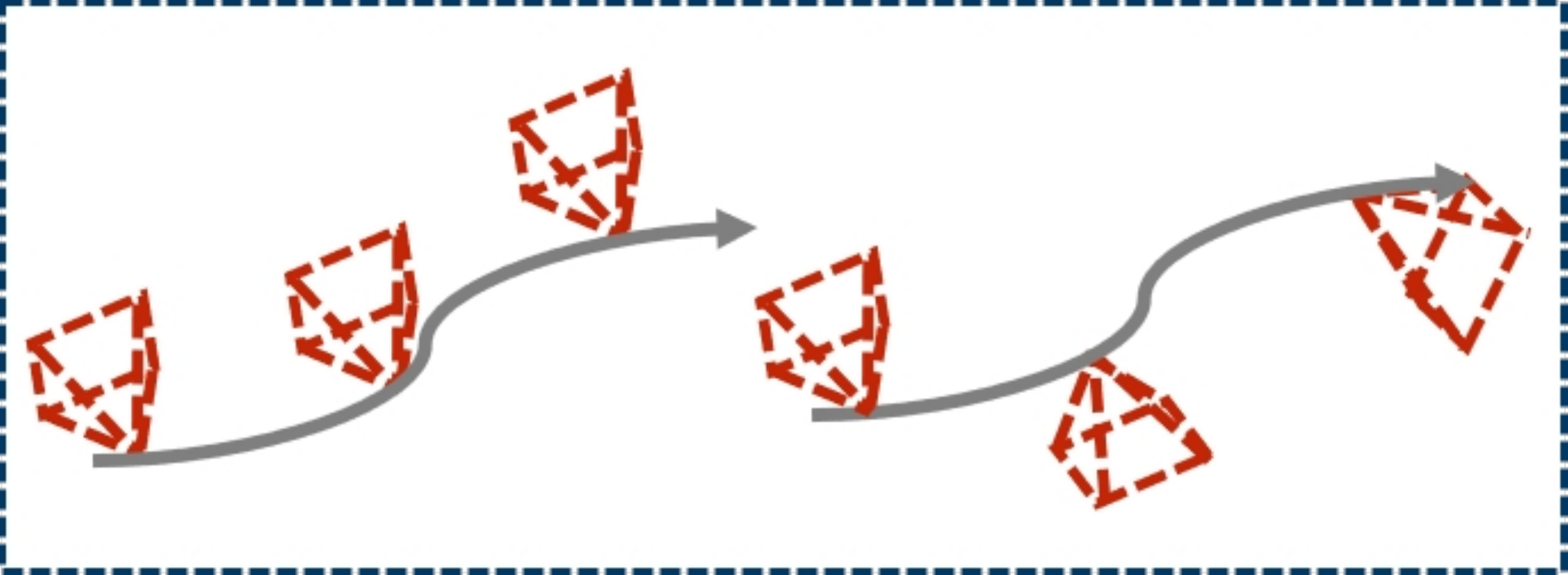}%
\label{fig2_1}}
\hfil
\subfloat[]{\includegraphics[width=0.45\linewidth]{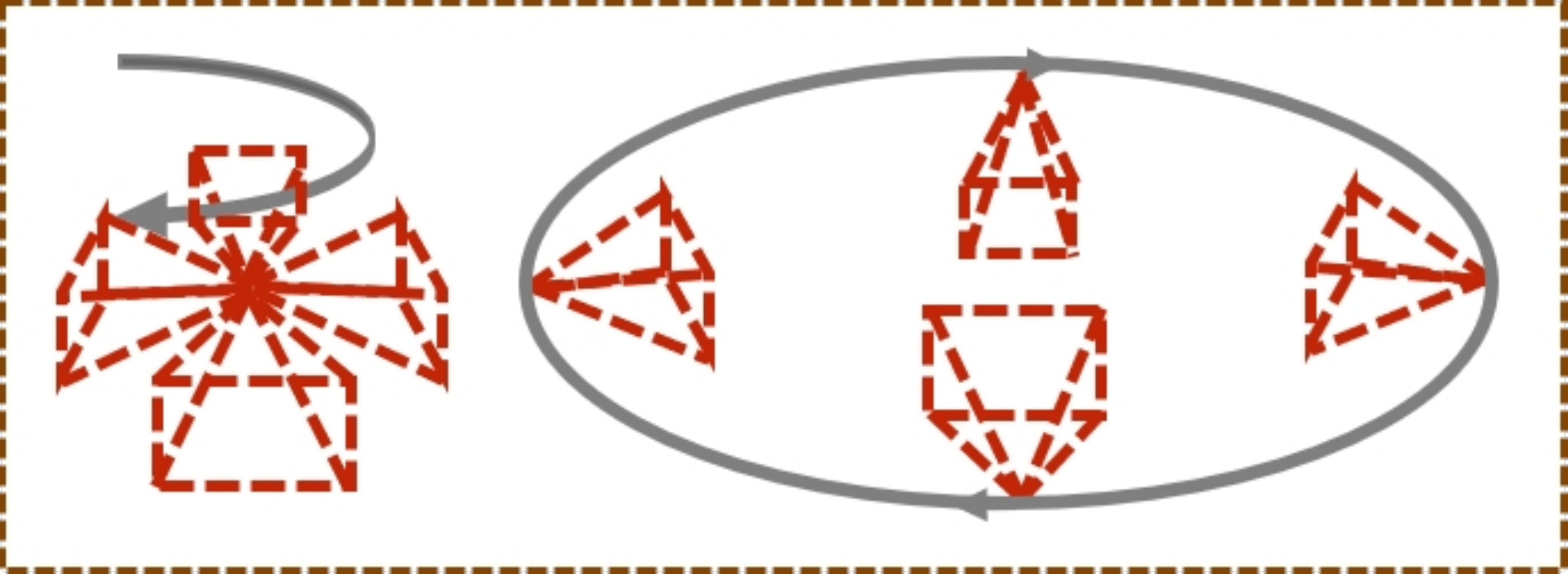}%
\label{fig2_2}}

\caption{Two cases of camera motion. The case of the same trajectory and different camera rotation is shown in (a). (b) denotes the opposite case.}
\label{fig2}
\end{figure}

\begin{figure}[!t]
\centering
\includegraphics[width=0.5\linewidth]{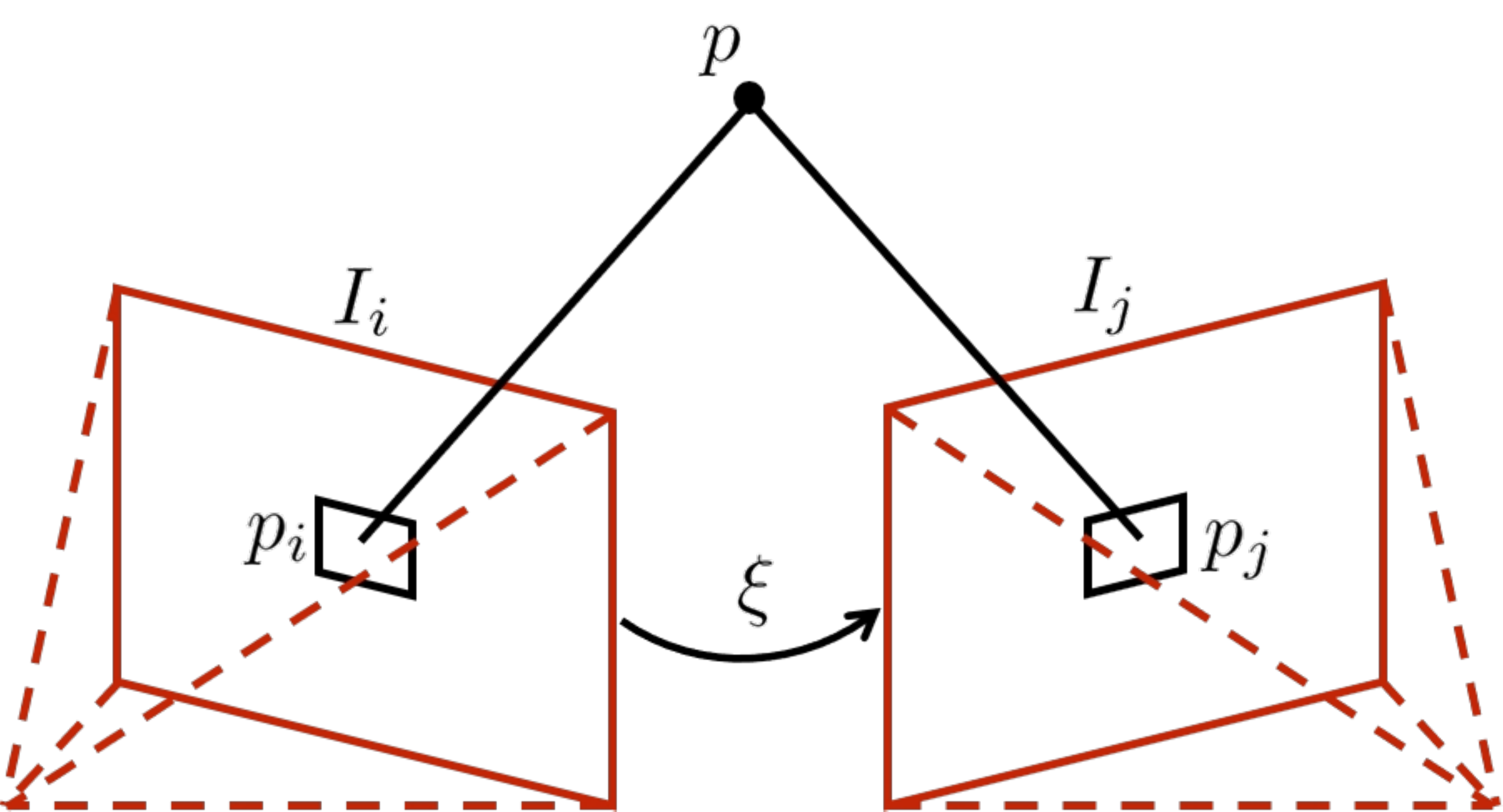}%

\caption{Camera pose estimation using a direct method.}
\label{fig3}
\end{figure}

While taking UAV videos, the trajectory of UAVs and movement of pan-and-tilt cameras both have an effect on the results of videos. Figure~\ref{fig2} shows a case with the same trajectory with different camera rotation and the opposite case. Evidently, flying the UAV and controlling the camera are of immense importance for air videography. Thus, we introduce a simultaneous localization and mapping (SLAM) framework proposed by~\cite{engel2018direct}, namely, direct sparse odometry (DSO), to estimate camera motion that can represent the motion feature of a video. DSO is a direct method that typically solves camera motion by minimizing photometric errors between pixels projected onto different frames from 3D points. As shown in Figure~\ref{fig3}, the problem can be formulized as:
\begin{equation}
\min_\xi J(\xi)=\mathop{\sum}_{j\in obs(p)}\|I_j(p_j)-I_i(p_i)\|^2,
\end{equation}
where $p_i$ and $p_j$ denote the 3D point $p$ observed in frame $I_i$ and $I_j$. $I(\cdot)$ means the observed pixel intensity in frame. $\xi$ denotes the camera pose, including translation and rotation. The camera pose and the spatial position of the 3D point, which are the inputs of the motion stream and structural stream, can finally be converged through iterative optimization.
Furthermore, photometric calibration of autoexposure videos~\cite{bergmann2018online} is applied to make the algorithm more robust to adapt to complex circumstances.

However, the shot length of videos is not constant, and only the keyframes are calculated for localization. It is assumed that the motion between keyframes is continuous, and this can be easily obtained due to the selecting principle of keyframes and the millisecond delay interval. Therefore, simple linear interpolation is applied to the three-dimensional vector, and spherical linear interpolation is applied to the quaternion. With respect to our problem, the expression is as follows:
\begin{equation}
t_{k}=\frac{\sin[(1-\frac{k}{n})\theta]}{\sin\theta}t_{m-1}+\frac{\sin(\frac{k}{n}\theta)}{\sin\theta}t_{m},
\end{equation}
where $n$ denotes the number of points that should be interpolated between $t_{m-1}$ and $t_{m}$, and $\theta$ denotes the central angle that can be calculated by $t_{m-1}$ and $t_{m}$.
All the vectors are normalized to avoid the limitation of all monocular SLAM, namely, scale ambiguity~\cite{strasdat2010scale}, which implies that the seven-dimensional vector can multiply any nonzero constant.

The 3D points that constitute the trajectories are not dense and unordered. In contrast, the relationship between adjacent points is still very close, and thus, we employ a network that differs from existing point cloud classification methods. The network of the motion stream is shown in Figure~\ref{fig1}. We use 1,024 points representing a trajectory as the input. Each point is represented by a seven-dimensional vector that consists of the UAV's position (XYZ) and the camera rotation quaternion.

In contrast to PointNet~\cite{qi2017pointnet}, we consider the relationship with neighboring points, and the first two convolution layers dealing with points are added. However, the dimension that denotes translation and rotation is not supposed to be convoluted. PointNet directly applies an affine transformation matrix to the coordinates of input points such that the learned representation by the pointset is invariant to geometric transformations. However, the input of our motion stream is the pointset representing camera motion, which is not supposed to be invariant to rigid transformation. For example, the crabbing trajectory can be obtained through a rigid body transformation from rectilinear flying, while they are two distinctly different types of camera motion. Thus, we design the motion stream network without any affine transformation matrix. Additionally, max pooling is not employed because some information related to camera motion may be lost.

It is also possible that the reconstructed structure of the scene affects UAV video aesthetics. It is controversial if only camera motion is considered. For example, fixed-point encircling is suitable for taking videos of towers, while rectilinear flying is more appropriate for shooting rivers. The reconstructed structure of the scene contains information on the shooting object size and scene layout. Thus, it is closely related to video aesthetics. Thus, we reconstruct the point cloud at the time when camera motion is estimated through the SLAM framework.

Subsequently, we used ConvNet to extract the structural features. The network of the structural stream is inspired by the PointNet architecture, which directly consumes point clouds and provides an approach for the object classification task. Specifically, we use the classification network of PointNet for our scene structure classification.

\subsection{Multistream Fusion}
Scene type is also considered by most individuals when determining whether a UAV video is professional. Occasionally, different rules of aesthetics are applied for different scene types. Thus, we optimize two branches with different learning objectives, including predicting aesthetic scores and classifying scene types in the bifurcated subnetwork. We connect the results computed from the three streams to two branches. The first branch is trained using binary labels to perform an aesthetic assessment. Additionally, the second branch is trained as a scene type classifier that is assumed as an improvement relative to the first branch.

Multiple feature fusion is an inevitable problem in most previous studies that use a multistream framework irrespective of whether the deep photoaesthetics assessment~\cite{mai2016composition} or deep networks for video classification are used~\cite{yang2013feature,wu2016multi}. The two-stream architecture for video classification is improved via several fusion methods such as max, sum, concatenate, and conv fusion~\cite{feichtenhofer2016convolutional}. Thus, we intend to apply a similar fusion strategy that is suitable for our task.

Given the prediction score $s^{k}$ of each stream $k (k=1, \ldots, N)$, the final prediction is as follows:
\begin{equation}
p=g(s^{1}, \ldots, s^{k}),
\end{equation}
where $g$ corresponds to a linear function or something else. Thus, we can fuse them by taking their average or considering them as features to an SVM classifier, which is termed late fusion \cite{simonyan2014two}. In this study, we perform an experiment using the averaging method. In the early fusion method, every stream is used as a feature extractor. Multimodal features from the three streams are extracted to determine the optimal fusion weights for each class. We can learn the optimal fusion weights via classifiers:
\begin{equation}
W=\mathop{\arg\min}_{w, b}-\frac{1}{m}\mathop{\sum}_{i=1}^m\{y_{i}\ln[\sigma(z)]+(1-y_{i})\ln[1-\sigma(z)]\},
\end{equation}
where $y$ denotes the ground-truth label, and $\sigma(z)$ where $z=wx+b$ denotes the actual output.

\section{Applications}

In this section, we present three application examples based on our proposed multistream framework, which proves that the UAV video quality assessment method can be applied to not only some common aesthetic evaluation tasks, such as helping video shooters and video sites to evaluate the quality of shooting but also more complex environmental exploration tasks, such as aesthetic-based path planning.
\subsection{UAV Video Grading}
\begin{figure}
\begin{center}
  \includegraphics[width=\linewidth]{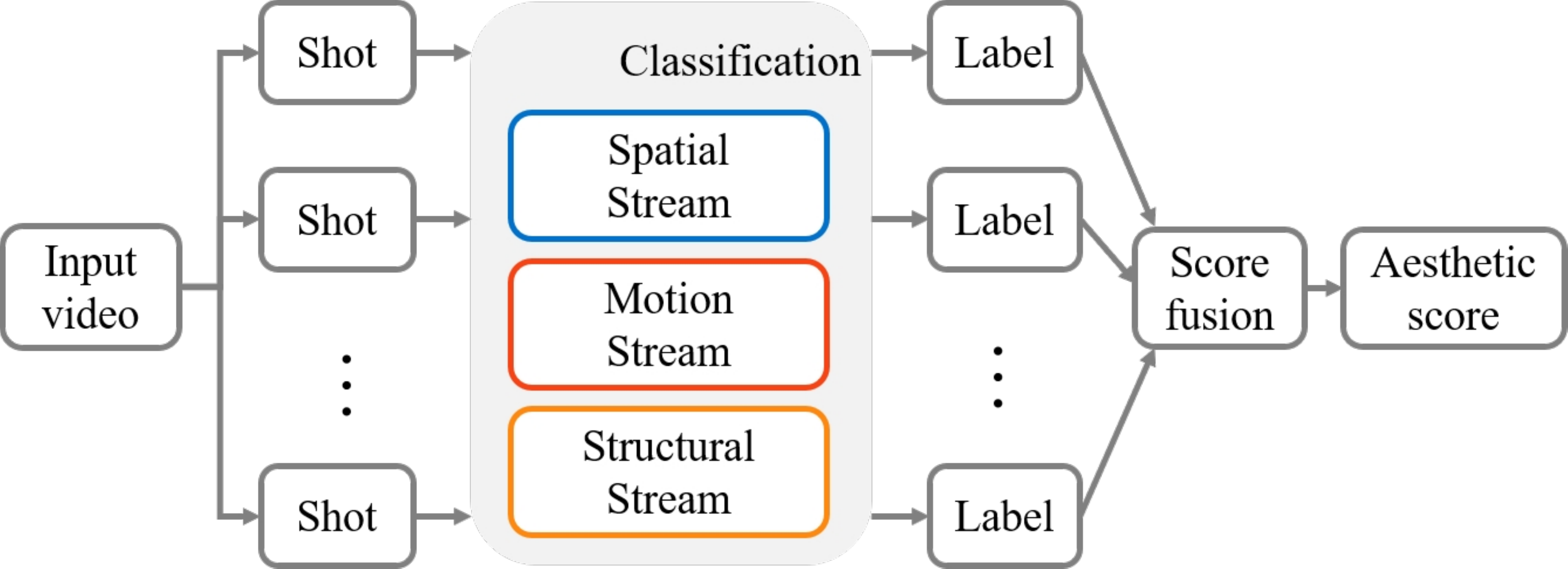}\\
\end{center}
  \caption{Grading UAV video overview. First, an input UAV video is segmented into several shots. Subsequently, our trained network is used as a classifier to obtain the aesthetic label of each shot. Finally, we consider all the labels to grade the input video.}\label{fig4}
\end{figure}

An application of our study involves automatically grading the UAV videos, including more than one shot, and this can be valuable to users or websites. Because videos with only one shot seem monotonous, most UAV videos are edited and consist of several shots. How to assess and grade videos with more than one shot is a noteworthy problem, especially for UAV video websites. It can also provide a reference for users to edit their videography works. Figure~\ref{fig4} shows how our proposed framework can be used. The video is initially divided into several shots by an automatic shot detection algorithm~\cite{apostolidis2014fast}. Subsequently, we can obtain the final aesthetic score based on the prediction probabilities of the shots. Because the shot length is not constant, the weighted average aesthetic score is calculated as the final score of the video. Thus, the final aesthetic score $a$ of a UAV video can be seen as:
\begin{equation}
a=\frac{a_{1}m_{1}+a_{2}m_{2}+\cdots+a_{n}m_{n}}{m_{1}+m_{2}+\cdots+m_{n}},
\end{equation}
where $a_{n}$ denotes the aesthetic score of shot $n$, and $m_{n}$ denotes the number of frames.

\subsection{Professional Segments Detection}
Another application involves detecting the professional segments of a UAV video. Amateur aerial videographers can be inquisitive as to how to take professional videos. A whole UAV flight can continue for approximately 15-40 min or longer. However, sometimes the segments that appear professional only last a few seconds. It may be difficult to choose the transitory professional segments in a lengthy video. Thus, we propose an application that can automatically detect fascinating segments with the network.

Given a segment length $m$, a UAV video can be cut into several segments $s_{1}, s_{2}, \cdots , s_{n}$. As mentioned above, camera motion $c$ and reconstructed point cloud $p$ can be simultaneously obtained. Thus, the goal involves obtaining the appropriate segment when the aesthetic score is maximum.
\begin{equation}
s=\mathop{\arg\max}_{s_{1}, \ldots, s_{n}}\{h(s_{1}, c_{1}, p_{1}), \cdots , h(s_{n}, c_{n}, p_{n})\},
\end{equation}
where $h$ denotes the prediction of our network, which is viewed as the probability of professional aerial shots.

\subsection{Aesthetic-based UAV Path Planning}
The above two applications relate to UAV videos. In addition, we also prove that our method is helpful for UAV path planning. Most previous studies for UAV path planning address obstacle avoidance~\cite{chen2013uav,iacono2018path,nguyen2018real} or minimum distance~\cite{zu2018multi}. However, few studies have focused on UAV video aesthetics. Thus, we present a modified A-star algorithm~\cite{yao2010path} for aesthetic-based UAV path planning. For each point $n$, the path score $F$ can be seen as:
\begin{equation}
F(n)=G(n)+H(n),
\end{equation}
where $G(n)$ represents the aesthetic score a drone achieves when it moves from the initial point to the current point $n$, $H(n)$ represents the predicted score the drone achieves when it moves from point $n$ to the termination point.
Unlike the traditional A-star algorithm, we intend to find the path that can obtain the highest aesthetic score instead of the shortest path.
\begin{figure*}[!t]
\centering
\subfloat[]{\includegraphics[width=\linewidth]{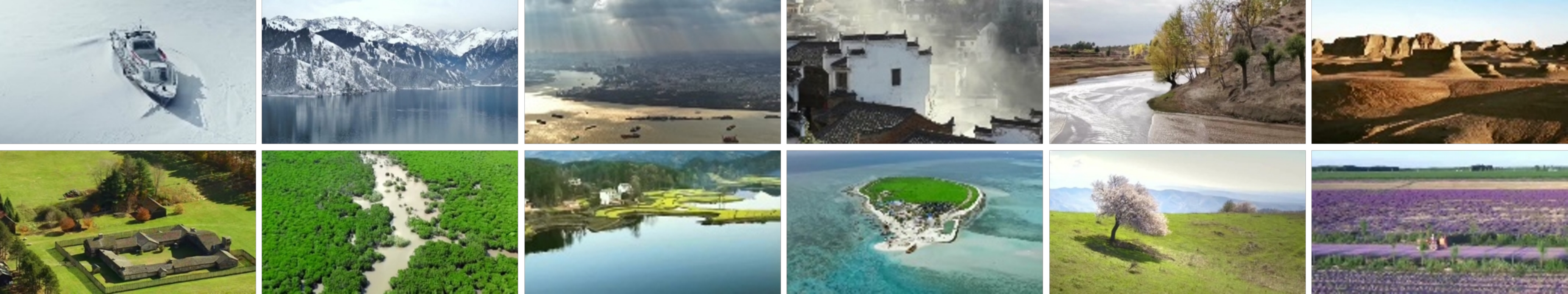}%
\label{fig5_1}}
\hfil
\subfloat[]{\includegraphics[width=\linewidth]{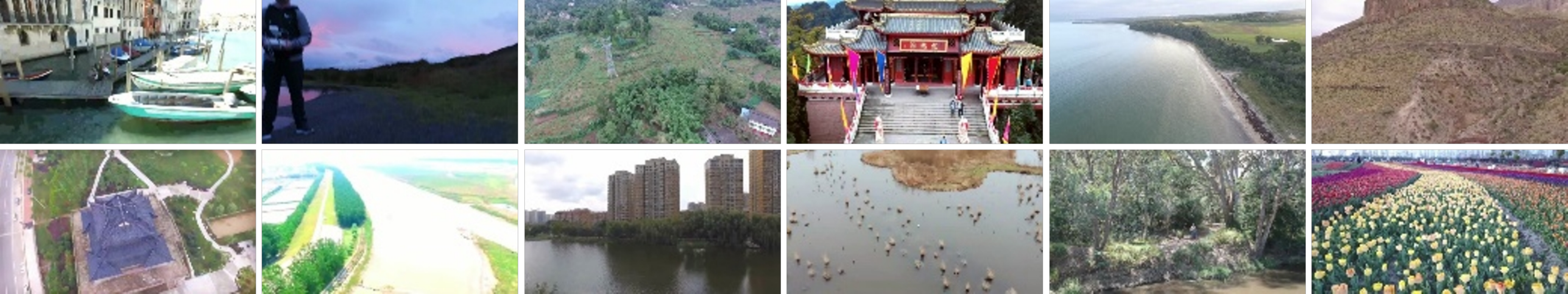}%
\label{fig5_2}}
\caption{First frames of the video sequences in our dataset. The professional UAV video shots are shown in (a), and (b) shows the amateur shots.}
\label{fig5}
\end{figure*}
\section{Dataset Statistics}
To the best of the authors' knowledge, there is a paucity of established datasets for UAV video aesthetic quality assessment. Hence, the aim of the study involves setting up a dataset including the videos made by experienced and amateur videographers that can be trained to explore the difference between them.

Inspired by previous studies in image and video aesthetics~\cite{tzelepis2016video,zhang2019gated}, we construct the \textbf{A}erial \textbf{V}ideo \textbf{A}esthetic \textbf{Q}uality (AVAQ) assessment dataset, which is utilized with the deep learning method.

We collected 6,000 UAV video shots, including 3,000 professional shots and 3,000 amateur shots, as shown in Figure~\ref{fig5}.

The professional shots are collected from several documentaries and films by considering that the documentaries and movies contain the connotative standards and rules of aesthetics and the experience of professional videographers and editors. These documentaries and films are highly rated on video sharing websites such as the Internet Movie Database (IMDb) and Douban, which encourage user-generated content. More than 30,000 people in total rate the videos with ratings ranging from one to ten, with ten indicating the most favorite video. The average score of these documentaries and films exceeds 9.0, which also shows that most people approve their professionalism.

Additionally, the amateur shots are downloaded from aerial video websites where amateurs can share their works with others. It should be noted that there is an abundance of UAV videos on the Internet. Some amateurs might share other fascinating aerial videos, including those aerial documentaries and good quality videos that professional videographers take. However, we noticed that the sites also recorded the devices used by the videographers. Therefore, to ensure the reasonability of our dataset, videos only taken using amateurs' own equipment are downloaded as opposed to their shared attractive aerial videos.

We focus on the aesthetic of frames and the camera trajectory as opposed to the shot change. Therefore, each video is an individual shot. Additionally, to avoid the influence of the shot length, the duration time of each shot does not exceed 1 min. The total length of AVAQ6000 is more than 25 hours. The resolution of videos is 720P or 1080P. The framerate is 30 fps.

\begin{figure}
\begin{center}
  \includegraphics[width=\linewidth]{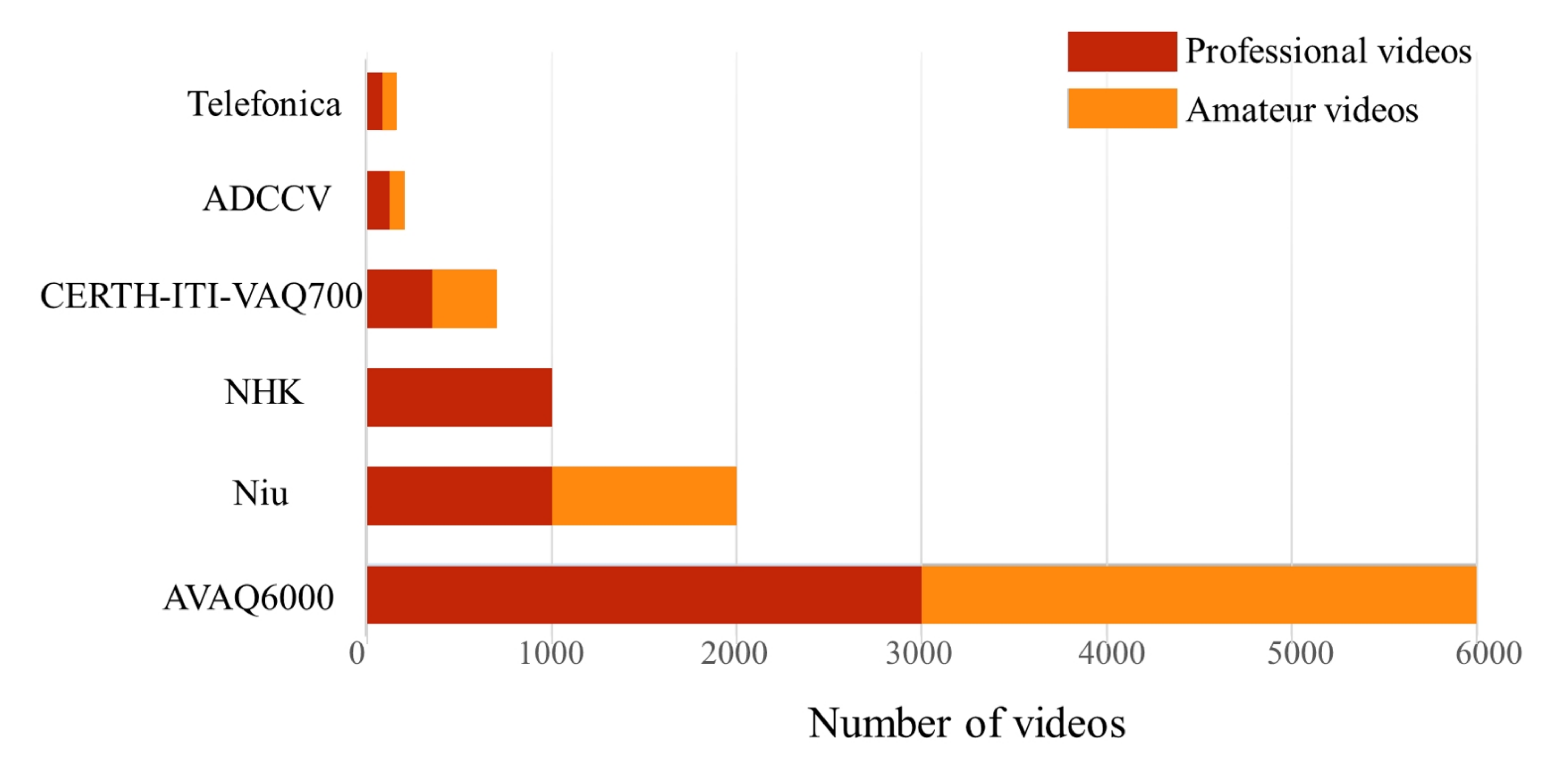}\\
\end{center}
  \caption{Datasets for video aesthetic quality assessment.}\label{fig6}
\end{figure}

In contrast to abundant image aesthetic quality assessment datasets, only a few video aesthetic quality assessment datasets are available. The existing datasets for video aesthetic quality assessment are listed in Figure~\ref{fig6}.

\begin{figure}
\begin{center}
  \includegraphics[width=\linewidth]{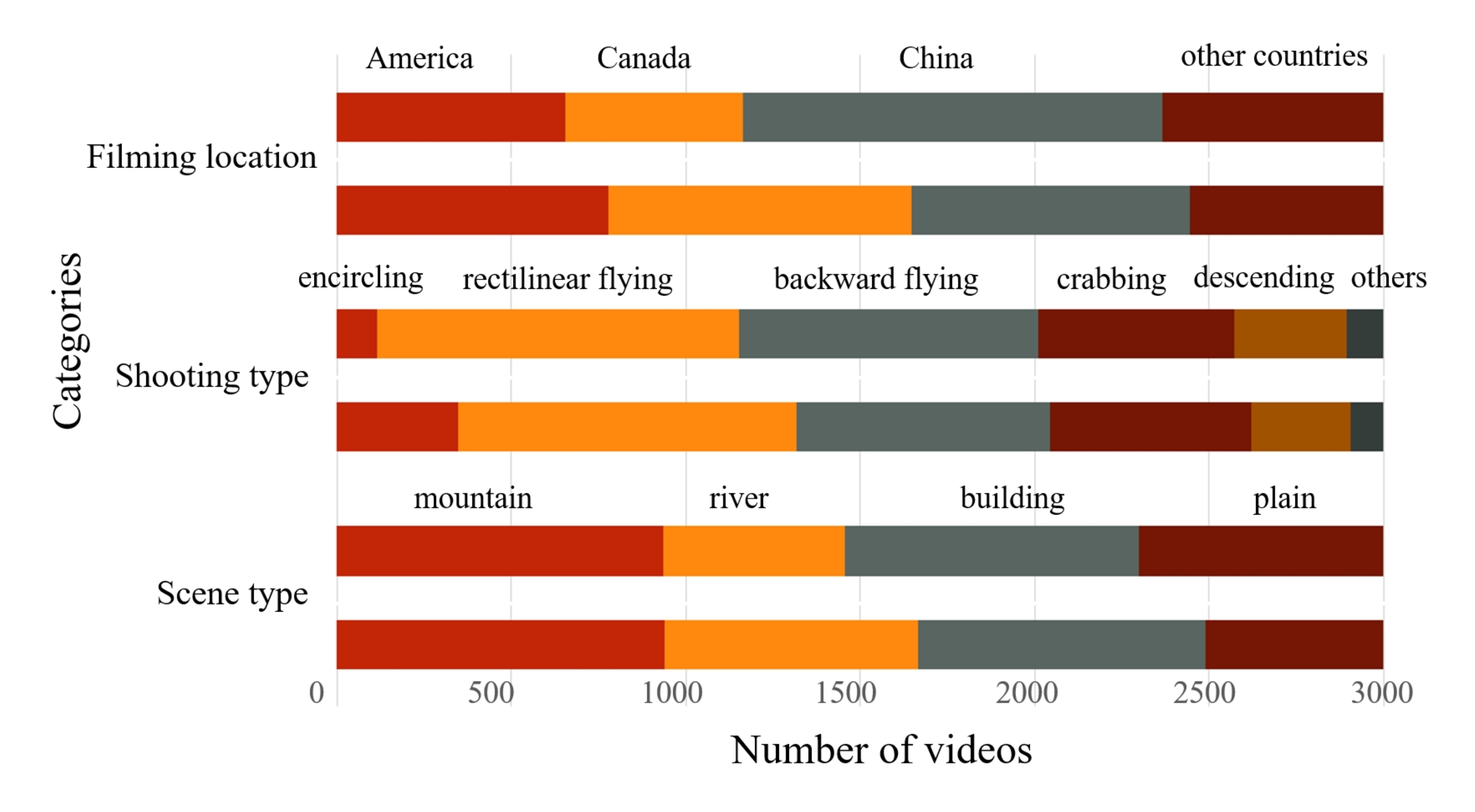}\\
\end{center}
  \caption{Statistics of our dataset. For each category, the first line denotes amateur UAV video shots, and the second line denotes professional shots.}\label{fig7}
\end{figure}

A good dataset should exhibit high diversity. Figure~\ref{fig7} shows how our dataset is constructed. We summarize the statistics of AVAQ6000 based on the following aspects:
\subsubsection{Filming location}
Professional video shots are collected from several documentaries and films that were shot in America, Canada, China, and other countries. Additionally, amateur video shots are downloaded from SkyPixel, YouTube, and other websites with keywords that are similar to those for professional videos.
\subsubsection{Shooting type}
Because AVAQ6000 is a dataset of aerial videos, the shooting type is an important element that differs from conventional video shooting. As the camera is mounted on the UAV, there are many shooting types relevant to the movement of the UAV.
Our dataset covers these different shooting types, including fixed-point encircling, rectilinear flying, backward flying, crabbing, and descending. We can also obtain some interesting conclusions from the statistics. It is easy to classify professional drone videos according to the method of shooting because experts usually use some fixed shooting techniques or a combination of methods. However, determining the categories of amateur drone videos is so difficult that we can only roughly classify them. It seems that some amateurs are not sure which shooting method should be used. Nevertheless, rectilinear flying is the most common method of shooting for both professionals and amateurs. This is likely because it is the most direct and convenient shooting method.
\subsubsection{Scene type}
We briefly group the video shots into four types: mountains, rivers, buildings, and plains. Sometimes multiple shooting scenes might appear in one video shot. In that case, we classify them according to the main content of the shooting videos. When we attempt to identify the scene type of a video, we find that the scene type is related to the method of shooting to some extent, which then affects the aesthetics of the drone video.

\section{Experiments}

\subsection{Spatial Stream}
As mentioned above, we define ConvNet with LSTM as our baseline. We initially extract all video frames. Given the fixed size of the network input, each video sequence is uniformly downsampled to 100 frames, which is the minimum length of all sequences. It should be noted that the captions or watermarks of the videos significantly affect the experiment. They can be viewed as nonnegligible noise that deviates from the expected results; thus, the final accuracy is unreasonable and unstable, and this can be excessively high or excessively low. Thus, the black edges, captions, and a few other things that are uncorrelated are cropped, such that it is only necessary to focus on the content.
\begin{table}
\caption{Comparison with other methods.}\label{table1}
\begin{center}
\begin{tabularx}{\linewidth}{c|X<{\centering} X<{\centering} X<{\centering} X<{\centering} X<{\centering}}
  \hline
  Method & Size (MB) & GFLOPs & Accuracy (\%) & F-score & AUC \\
  \hline
  C3D \cite{tran2015learning} & 390 & 0.16 & 68.75 & 0.72 & 0.77 \\
  CNN-GRU & 22 & 0.01 & 66.62 & 0.71 & 0.75 \\
  Inception V3 + LSTM & 365 & 0.03 & 74.08 & 0.76 & 0.83 \\
  ResNet V2 + MLP & 366 & 0.06 & 75.78 & 0.78 & 0.84 \\
  ResNet V2 + LSTM & 365 & 0.03 & {\bf 77.31} &{\bf 0.80} & {\bf 0.86} \\
  \hline
  SVM-based~\cite{tzelepis2016video}& ----- & ----- & 64.00 & ----- & ----- \\
  Baseline & ----- & ----- & 69.87 & ----- & ----- \\
  \hline
\end{tabularx}
\end{center}

\end{table}
We compare the spatial stream with two other common video classification methods, C3D~\cite{tran2015learning} and CNN-GRU, as shown in Table~\ref{table1}. Additionally, we also compare two networks, namely, Inception V3 and ResNet V2, which perform significantly well on ImageNet for local feature extraction. The top layers of both networks are replaced by the same modified fully connected layers or LSTM layers based on our task, notwithstanding the fact that Inception V3's features include 2,048 dimensions, while the latter's features include 1,536 dimensions. The results are shown in Table~\ref{table1}. We also present the model sizes and computational complexity of different models.

The LSTM layer with ResNet V2 as the feature extractor exhibits the best performance.
The results indicate that our baseline performs better when compared to several traditional methods for video classification.

Furthermore, we present comparative results relative to the traditional SVM-based video quality assessment method. Only a few datasets for video aesthetic quality assessment are publicly available and sufficiently abundant for deep learning; thus, we can only use the spatial stream as the feature extractor and use the features as the input of traditional SVM. The strategy is applied for comparative experiments in a previous study on image aesthetic assessment.

Our experiment involves CERTH-ITI-VAQ700, which is the only available dataset for video aesthetic assessment, to the best of our knowledge. Specifically, 350 professional videos and 350 amateur videos are extracted via the spatial stream to obtain high-dimensional features. Subsequently, KSVM is applied for classifying as~\cite{tzelepis2016video}. Table~\ref{table1} shows that the spatial stream performs better than the traditional SVM-based method that uses handcrafted aesthetic features as inputs.


\begin{table}
\caption{Effect of translation and rotation.}\label{table2}
\begin{center}
\begin{tabularx}{\linewidth}{c|X<{\centering} X<{\centering} X<{\centering} X<{\centering} X<{\centering}}
  \hline
  Method & Size (MB)& GFLOPs& Accuracy (\%) & F-score & AUC\\
  \hline
  Translation & 67 & 0.03 & 69.12 & 0.70 & 0.75 \\
  Rotation & 67 & 0.03 & 71.59 & 0.73 & 0.81 \\
  T\&R (PointNet)~\cite{qi2017pointnet}& 14 & 0.01 &76.46 & 0.78 & 0.85\\
  T\&R (Ours) & 67 & 0.03& {\bf77.85} & {\bf0.78} & {\bf0.86} \\
  \hline
\end{tabularx}
\end{center}

\end{table}

\subsection{Motion Stream and Structural Stream}
The frames of UAV videos are photometrically calibrated prior to estimating camera motion to make the algorithm more robust. Additionally, the results of our experiment indicate that photometric calibration significantly aids in a few extreme conditions, including sunset or night, when illumination significantly changes, which violates the assumption of the direct method of SLAM wherein the illumination is constant. Subsequently, we use DSO to estimate camera motion with the calibrated frames. The camera motion consists of a sequence of points. Additionally, each point is represented by a 7-dim vector of XYZ and a quaternion. With respect to each sequence of points, the number of points is interpolated to 1,024 based on different video durations.
\begin{table*}
\caption{Results for the individual stream and multi-stream.}\label{table3}
\begin{center}
\begin{tabularx}{0.95\linewidth}{c|X<{\centering} X<{\centering}|X<{\centering} X<{\centering} X<{\centering}|X<{\centering} X<{\centering} X<{\centering}}
  \hline
  \multirow{2}{*}{Method}
  & \multicolumn{2}{c|}{Model}& \multicolumn{3}{c|}{Task1 (aesthetic label)}& \multicolumn{3}{c}{Task2 (scene type)}\\
  \cline{2-9}
  &Size (MB)&GFLOPs& Accuracy (\%)& F-score & AUC & Accuracy (\%) & F-score & AUC\\
  \hline\
  Spatial stream& 365& 0.03& 78.74 & 0.80 & 0.85 & 75.13 & 0.49 & 0.66 \\
  Motion stream & 68& 0.03&  78.02 & 0.79 & 0.84 & 37.89 & 0.28 & 0.54 \\
  Structural stream& 14 & 0.01&  67.52 & 0.72 & 0.73 & 35.58 & 0.26 & 0.51 \\
  \hline\
  Spatial \& Motion& 408 & 0.06 & 86.21 & 0.85 & 0.91 & 76.04 & 0.50 & 0.67 \\
  Spatial \& Structural& 376 & 0.04&  79.91 & 0.81 & 0.86 & 75.26 & 0.49 & 0.66 \\
  Motion \& Structural& 88 & 0.03&  79.73 & 0.81 & 0.89 & 37.92 & 0.28 & 0.55 \\
  \hline\
  Multistream (late fusion) &418 & 0.07&  87.84 & 0.87 & 0.92 & 77.44 & 0.52 & 0.70 \\
  Multistream (early fusion) &417 & 0.07&  {\bf 89.12} & {\bf 0.88} & {\bf 0.95} & {\bf 78.62} & {\bf 0.53} & {\bf 0.71} \\
  \hline
\end{tabularx}
\end{center}

\end{table*}

To prove that both translation and rotation are relevant to video aesthetic quality, we perform individual experiments on them. First, we use the estimated UAV trajectories of AVAQ6000 as the input. We only select the three-dimensional vector that represents the trajectory of the UAV to test the translation. For the rotation experiment, we choose the quaternion, which represents the rotation of the camera mounted on the UAV as the input. Then, we train the motion stream, which is shown in Figure~\ref{fig1}. However, as we only experiment on them individually, the output of the network is a single value.

Table~\ref{table2} shows the results of the motion stream. It can be seen that methods combining translation and rotation achieve better results, and this demonstrates that our approach that combines the flight path of UAVs and movement of mounted cameras aids in UAV video aesthetic quality assessment. We also compare our motion stream network with another point cloud classification method, PointNet. We consider that the relationship between 3D points cannot be ignored because of the ordered and sparse points that constitute the trajectories. Thus, the motion stream network adds convolution layers dealing with points instead of the max pooling layer that PointNet employed. The results prove that our motion stream network is more appropriate for 3D trajectory classification.

The sparse 3D point cloud can also be reestablished while estimating camera motion. However, the task for monocular video sequences is more difficult than that for multiview sequences. Because of the uncertainty of estimated points, the reconstructed point clouds are intermixed with a number of unexpected points. Therefore, we preprocess the point cloud by denoising the point cloud. The points that are outliers can be seen as the noise that should be discarded. Subsequently, we sample 4,096 points via a voxel grid filter that can maintain the structural characteristic of the reconstructed scene.

\subsection{Multistream Fusion}

Finally, we experiment on the multistream network with two different fusion methods. First, the late fusion strategy is applied by calculating the average value of the three predicted results as the final aesthetic label and the prediction of scene type. As shown in Table~\ref{table3}, it obtains a better result when compared with each individual stream.

Subsequently, the early fusion ploy is put into effect via merging the multimodal features that consist of photoaesthetics, camera motion, and scene structure features at the second to last fully connected layer. Three more fully connected layers are added before the final results. As shown in the table, the accuracy exceeds the late fusion, which demonstrates that early fusion that learns the best fusion weights performs better than the simple averaging method. It should be noted that the results of task 2, which classifies the scene type in motion and structural streams, are not as good as in the spatial stream because the scene type, including building, mountain, river and plain, is defined according to the content of frames. Thus, the scene type classification task is more relevant to the spatial stream. However, the results of the multistream network also reveal that motion and structure streams are beneficial to scene type prediction.

\begin{figure*}[!t]
\centering
\subfloat[]{\includegraphics[width=0.95\linewidth]{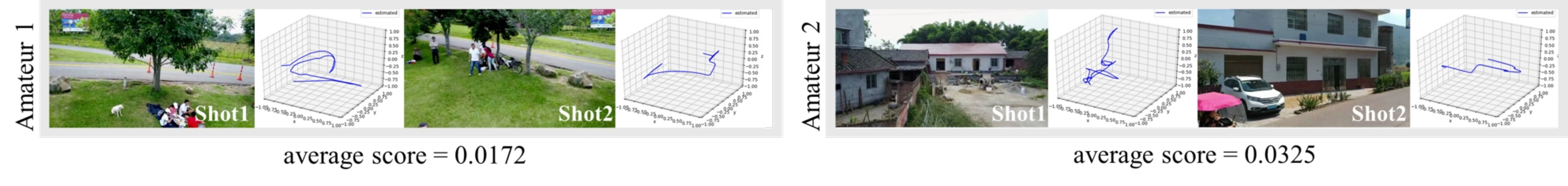}%
\label{fig8_1}}
\hfil

\subfloat[]{\includegraphics[width=0.95\linewidth]{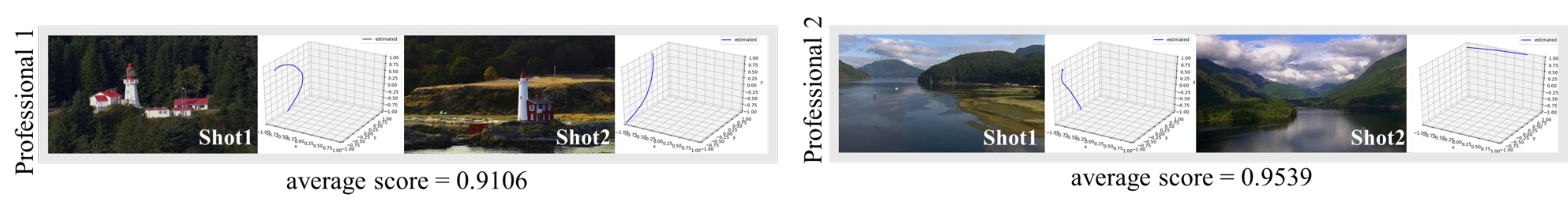}%
\label{fig8_2}}
\caption{(a) shows two amateur videos with the lowest scores and (b) presents the professional videos with the highest scores.}
\label{fig8}
\end{figure*}

\subsection{Implementation Details}
At the training time, we randomly choose 4,200 shots, which approximately corresponds to 70\% of AVAQ6000. Additionally, we assign 10\% as a validation set and 20\% as a test set. To avoid the effect of data selection, we repeat 5 times and obtain the mean values as the results in all our experiments.

When we train the individual streams, we do not set the same learning rate and decay due to the differences in network architecture. With respect to the spatial stream, the parameters of the layers before the first fully connected layer of the pretrained GoogLeNet models, including Inception V3 and ResNet V2, are fixed for fine tuning. The stochastic gradient descent is used to train our model with a learning rate of 0.001 and a weight decay of 0.0001. Additionally, in the motion and structural stream, we set the learning rate as 0.01 and the decay as 0.01.

At early fusion, the sizes of three more fully connected layers are 512, 256 and 6. To facilitate the training process of the multistream network, we transfer the parameters of each stream and retrain them with a very small learning rate. The training time approximately corresponds to 1 d using GTX1080-Ti GPU.

\subsection{Applications}

Here, we present three application examples.
\subsubsection{UAV Video Grading}
First, we collect 20 UAV videos on the Internet that are not included in our dataset. Like most videography works, these videos consist of more than one shot (approximately 5-10 shots). Subsequently, they are segmented into several shots automatically~\cite{apostolidis2014fast}. As the shot frames and the trajectory points are sampled in the preprocessing stage, shots with variational lengths are handled. Then, shots are graded based on their aesthetic scores as predicted by our network. Finally, we calculate the weighted average of the scores as the aesthetic score of the entire video.

Figure~\ref{fig8} shows two videos with the highest and two videos with the lowest scores among the 20 collected UAV videos. The first frame and the camera motion of two shots of each video are presented in the figure. Evidently, videos with fascinating shots obtain higher scores than those with unattractive shots. The results indicate that our method can differentiate between professional and amateur UAV videos.
\subsubsection{Professional Segments Detection}
The second application involves detecting the professional segments of UAV videos taken by amateur users, and this can provide an effective method for users to edit their aerial videography works without a significant amount of specialized knowledge.

\begin{figure}
\begin{center}
  \includegraphics[width=0.8\linewidth]{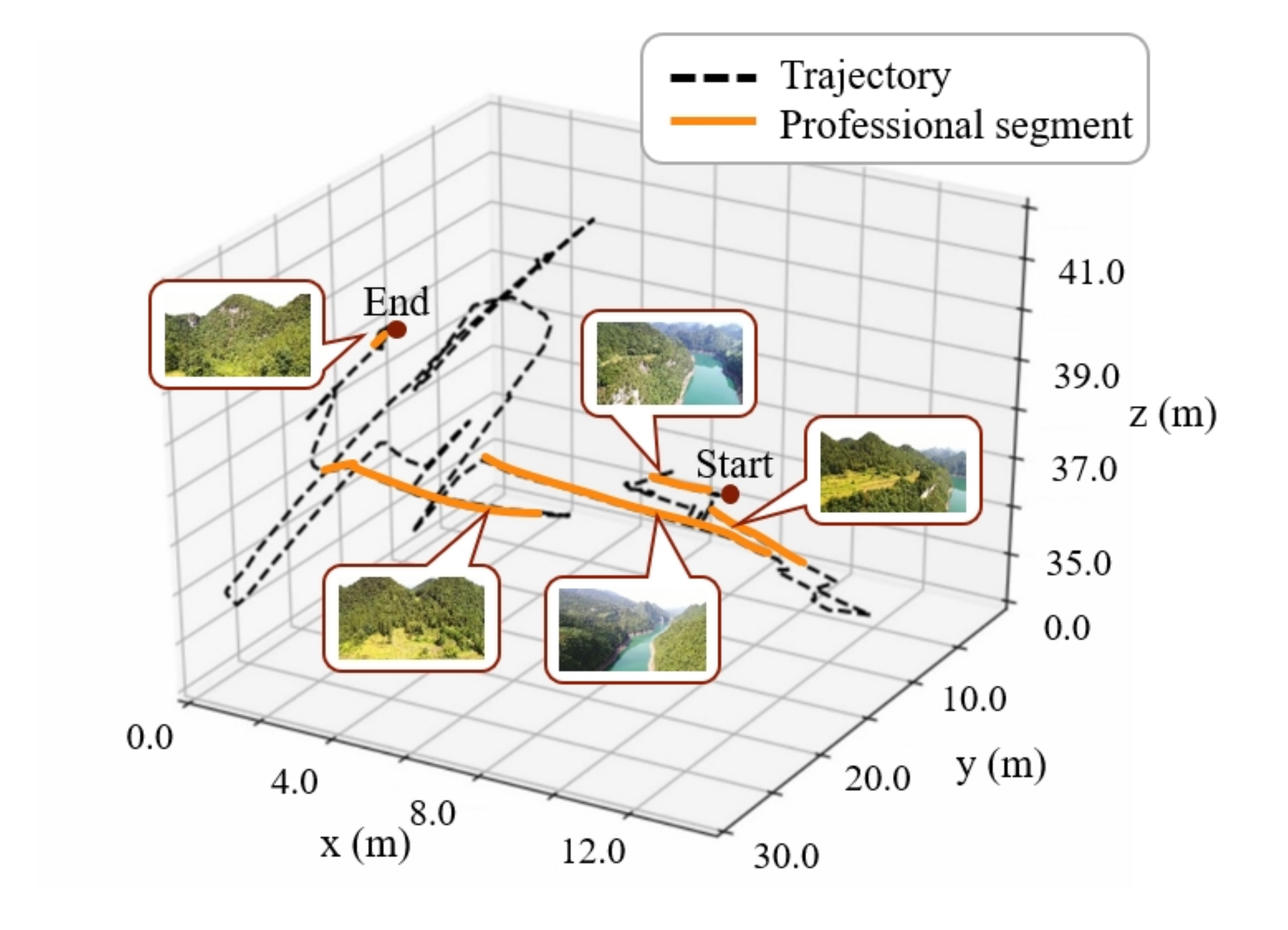}\\
\end{center}
\caption{Professional aerial video segment detecting.}\label{fig9}
\end{figure}

We select a common video taken by an amateur user that lasts for more than 10 min. It includes a complete flight including taking off and landing. We set the segment length as 300 frames because this is frequently used as a shot length in professional documentaries or movies. Subsequently, the network predicts the aesthetic label such that the professional segments can be elected. The results are shown in Figure~\ref{fig9}.

\subsubsection{Aesthetic-based UAV Path Planning}
In addition, we present how our method can generate an aesthetic-based UAV path. First, we fly a drone and shoot the scene casually, similar to photographers attempting to find the best view. Then, we utilize Altizure, a 3D modeling software, to model the realistic scene using several aerial photos so that we could calculate a desired path avoiding the obstacles. Given the initial point and the termination point, the A-star algorithm calculates the appropriate waypoints step-by-step. For our task, we find the waypoints that obtain the highest predicted aesthetic score instead of the shortest Manhattan distance. Specifically, we sample the waypoints every 2 meters in the up and down, right and left, front and back directions. Considering that camera motion is very important for video aesthetics, we set the yaw angle $\pm 5$ or $0$ when each waypoint is calculated. In other words, we calculate 18 possible waypoints at each step. Because we can obtain the aerial video and the trajectory in the virtual scene, it is possible to obtain the aesthetic score by our network.
Then, we use the minimum snap trajectory algorithm~\cite{mellinger2011minimum} to generate a smooth trajectory that the drone can follow. Finally, we use DJIWaypointMission SDK to allow the drone to shoot in the real world. For comparison, we set the speed to 1 meter per second in all cases.
\begin{figure*}[!t]
\centering
\subfloat[]{\includegraphics[width=0.95\linewidth]{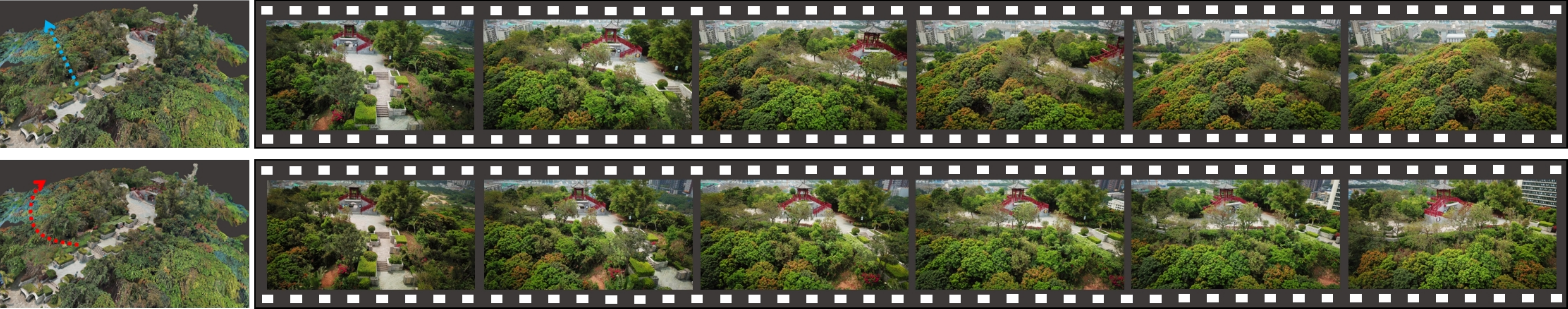}%
\label{fig11_1}}
\hfil
\subfloat[]{\includegraphics[width=0.95\linewidth]{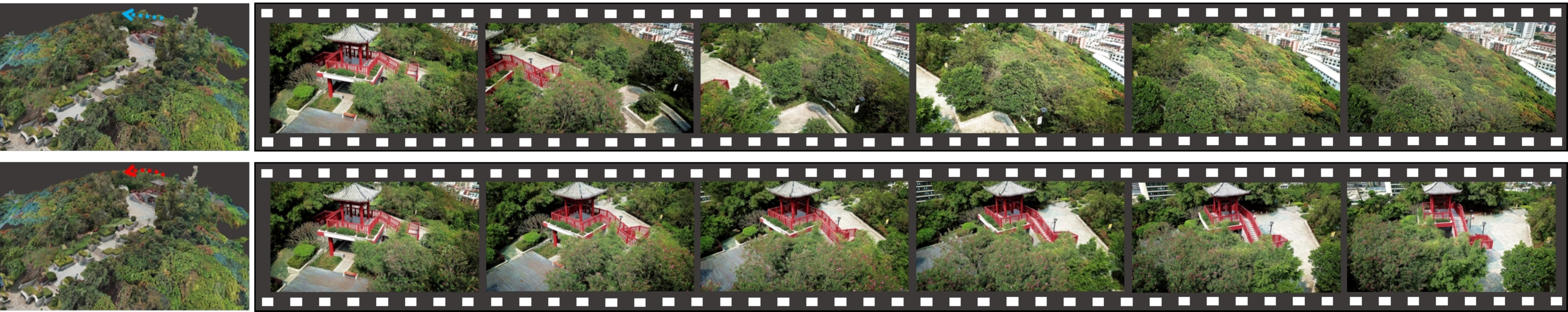}%
\label{fig11_2}}
\hfil
\subfloat[]{\includegraphics[width=0.95\linewidth]{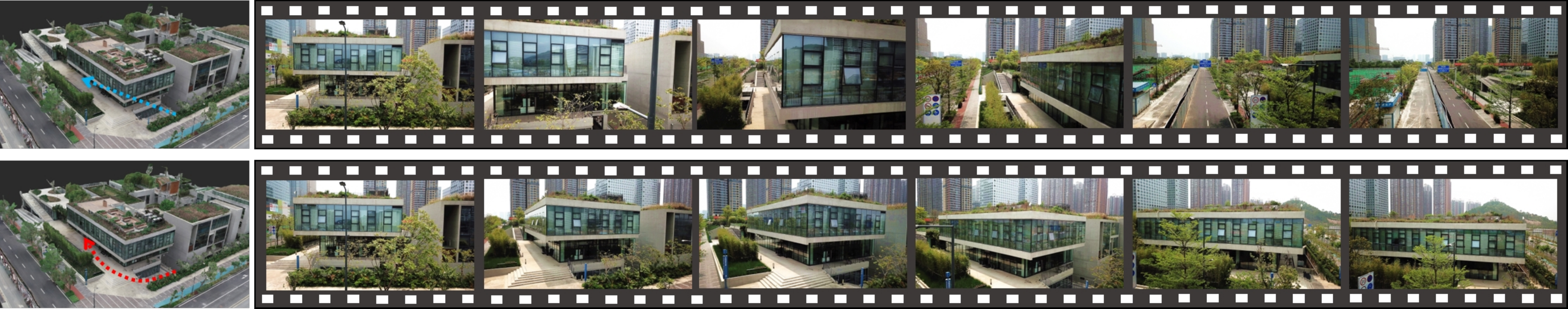}%
\label{fig11_3}}
\hfil
\subfloat[]{\includegraphics[width=0.95\linewidth]{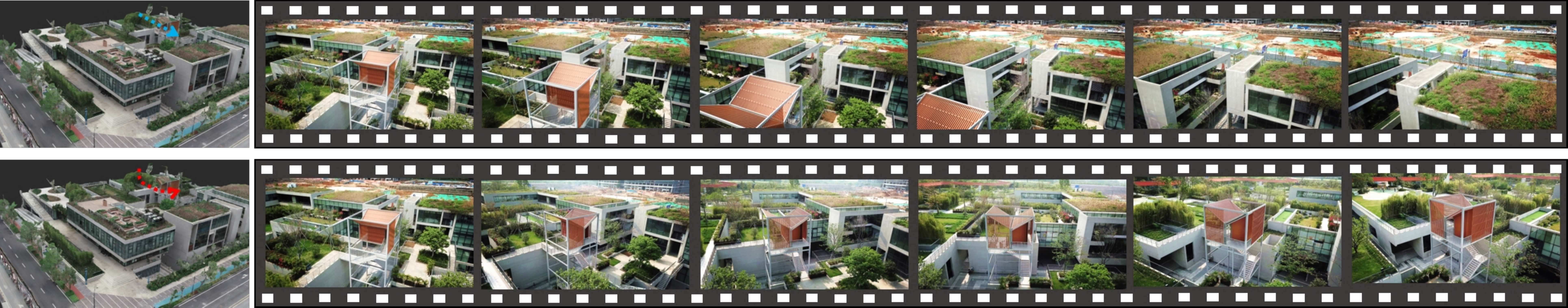}%
\label{fig11_4}}
\caption{Comparison with the A-star algorithm. The first picture of each row presents the reconstructed 3D scene. The first row shows the A-star path in blue, and the second row shows the aesthetic-based path in red.}
\label{fig11}
\end{figure*}

\begin{figure}[!t]
\centering
\subfloat[]{\includegraphics[width=0.45\linewidth]{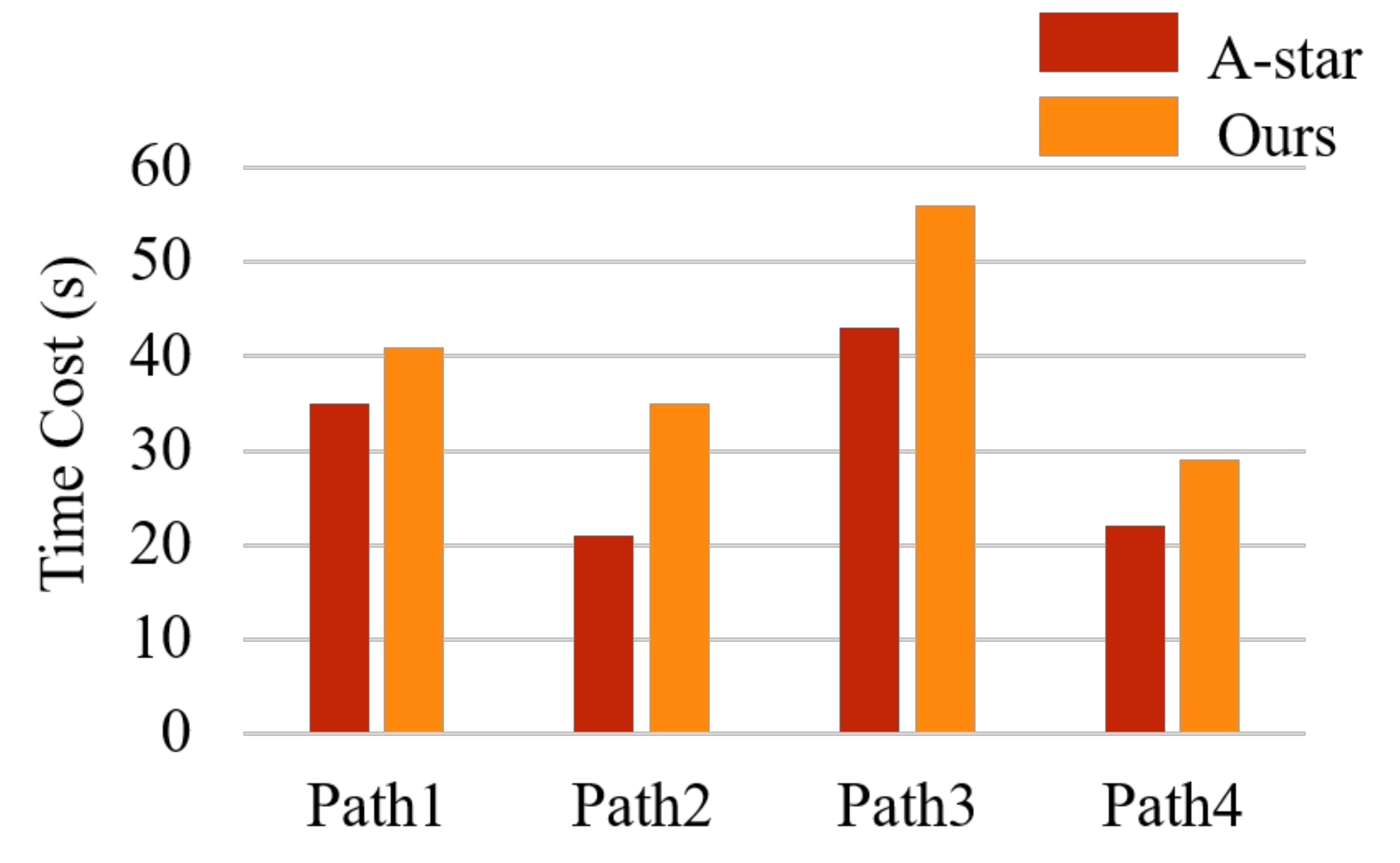}%
\label{fig10_1}}
\hfil
\subfloat[]{\includegraphics[width=0.45\linewidth]{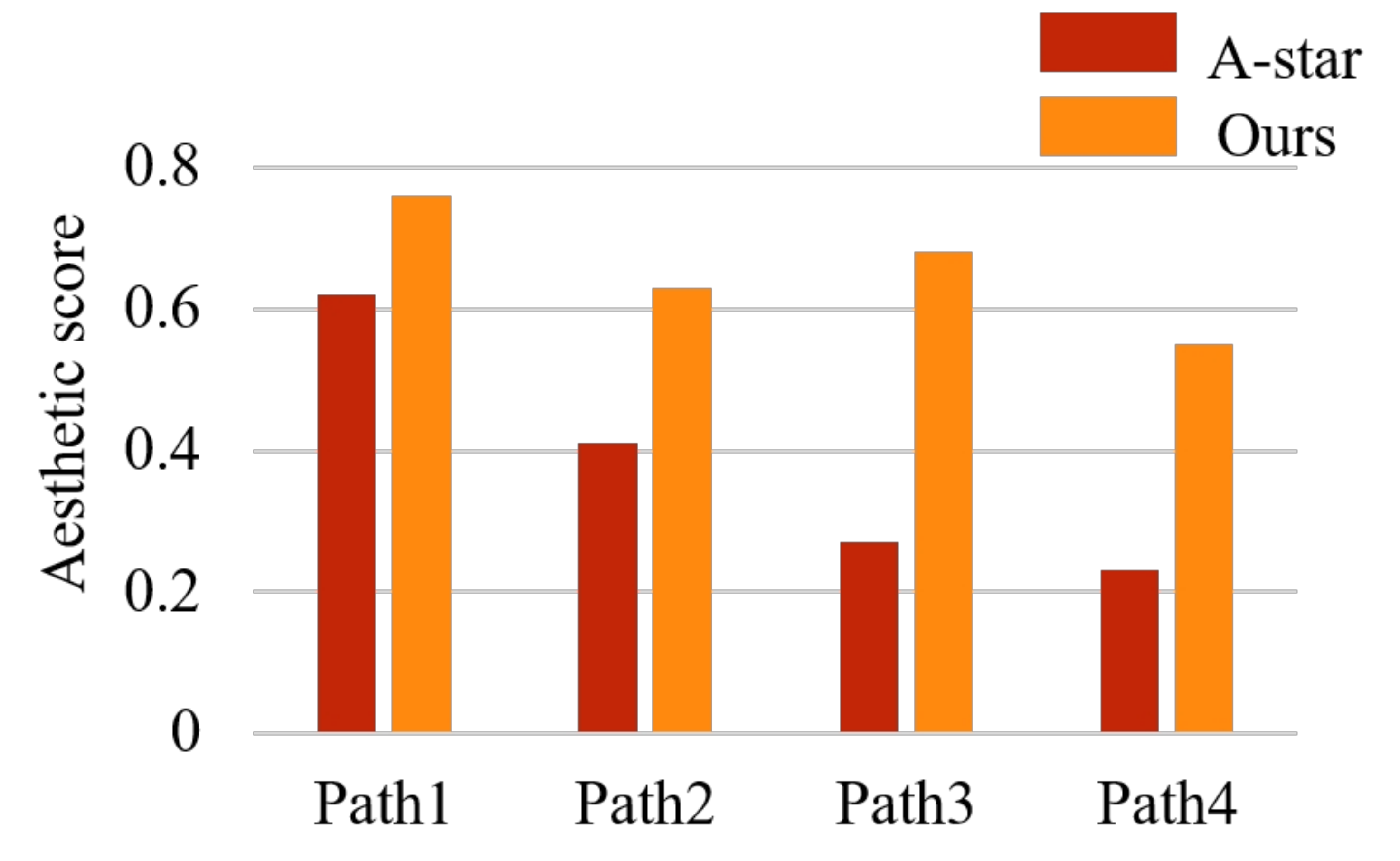}%
\label{fig10_2}}
\caption{Time cost and predicted aesthetic score comparison.}
\label{fig10}
\end{figure}

Here, we compare our aesthetic-based UAV path planning method with the traditional A-star algorithm. We test them in two scenarios. In each scenario, we select two different sets of starting points and ending points.
Fig~\ref{fig11} shows the footage that the drone shot with different methods in the real world.
We compare their time cost and the aesthetic score predicted by our method, which is shown in Fig~\ref{fig10}. Although it seems that our method consumes 30\% more time than the traditional A-star algorithm on average, we obtain a more attractive aerial video.

\section{Conclusions and Future Work}
In this study, we presented a method of deep multimodality learning for UAV video aesthetic quality assessment. A multistream network combining spatial, motion and structural streams was proposed for exploiting multimodal clues, including spatial appearance, drone camera motion, and scene structure. We constructed a dataset containing 6,000 UAV video shots, and this is the first dataset for UAV video aesthetics to the best of the authors' knowledge. Additionally, we designed a novel network to maximize the relationship between neighboring track points for exploring the characteristics of 3D trajectories. The results indicated that our method can learn the aesthetic features of UAV videos to distinguish between professional and amateur videography works. In addition, we presented three application examples to prove that our method is practical and has important implications.

However, although the proposed network is effective, it has a few limitations. We use a SLAM framework to estimate camera motion. Hence, a few cases of failure exist due to the algorithm. Additionally, our network is not end-to-end. Therefore, one interesting future work will replace the SLAM framework with a neural network to make the method more stable and applicable. The ConvNet specialized for video aesthetics is worthwhile. Moreover, we will expand the quantity and improve the quality of our dataset.

\appendices



\ifCLASSOPTIONcaptionsoff
  \newpage
\fi



%
\bibliographystyle{IEEEtran}
\bibliography{IEEEabrv,egbib}




%
\begin{IEEEbiography}
[{\includegraphics[width=1in,height=1.25in,clip,keepaspectratio]{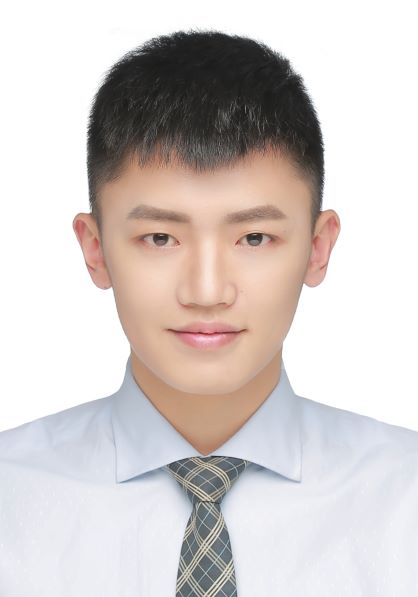}}]
{Qi Kuang} received his B.S. degree in Computer Science from Beihang University, China, in 2015. He is currently pursuing the Ph.D. degree in Technology of Computer Application from Beihang University. His research interests include Computer Graphics and Robotics.
\end{IEEEbiography}

\begin{IEEEbiography}
[{\includegraphics[width=1in,height=1.25in,clip,keepaspectratio]{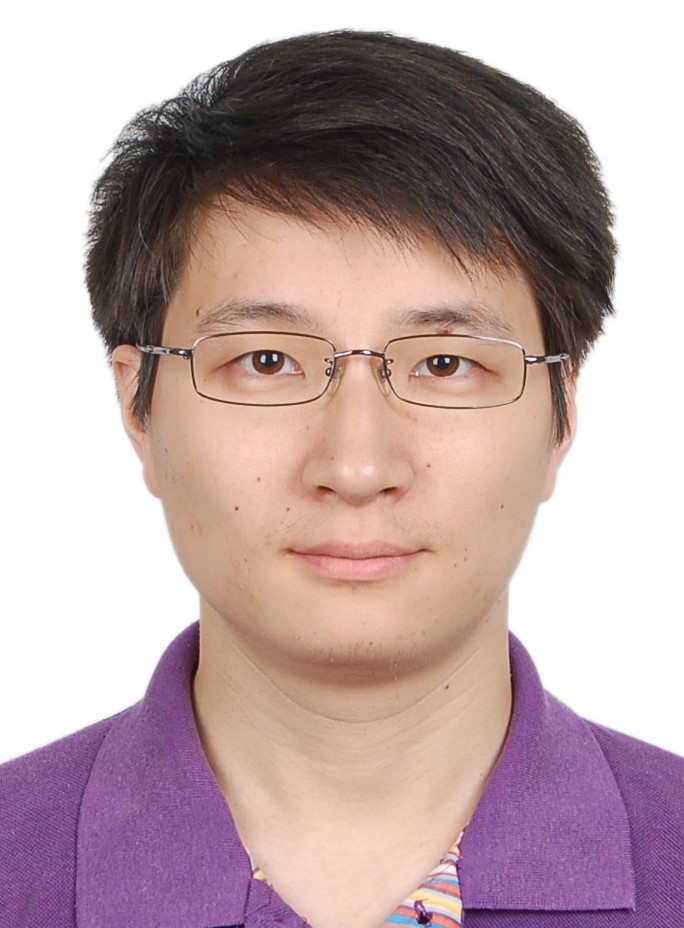}}]{Xin Jin}
received his Ph.D. degree in Technology of Computer Application from Beihang University, China, in 2013. He is currently an Associate Professor with the Department of Cyber Security, Beijing Electronic Science and Technology Institute. His research interests include Visual Computing and Visual Media Security.
\end{IEEEbiography}
\vspace{-110 mm}
\begin{IEEEbiography}[{\includegraphics[width=1in,height=1.25in,clip,keepaspectratio]{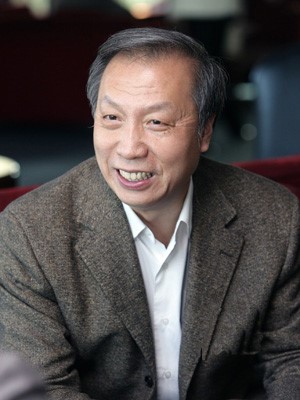}}]{Qinping Zhao}
received his Ph.D. degree in Computer Science from Nanjing University, China, in 1986. He is currently a Professor with the State Key Laboratory of Virtual Reality Technology and Systems, School of Computer Science and Engineering, Beihang University. He is the Fellow of Chinese Academy of Engineering, and the Founder of State Key Laboratory of Virtual Reality Technology and Systems. His research interests include Virtual Reality and Artificial Intelligence.
\end{IEEEbiography}
\vspace{-110 mm}
\begin{IEEEbiography}[{\includegraphics[width=1in,height=1.25in,clip,keepaspectratio]{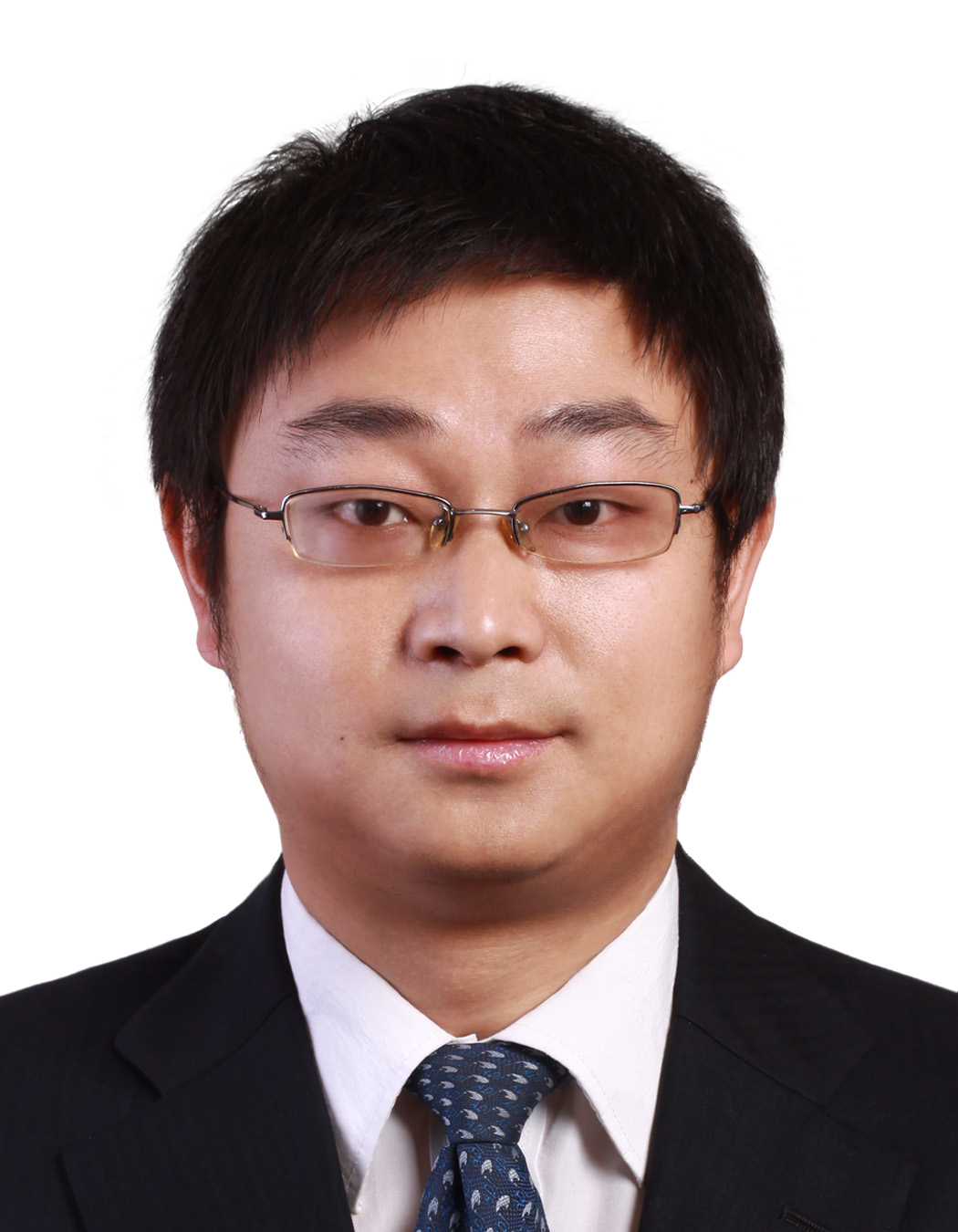}}]{Bin Zhou}
received his B.S. and Ph.D. degrees in Computer Science from Beihang University, China, in 2006 and 2014, respectively. He is currently an Assistant Professor with the State Key Laboratory of Virtual Reality Technology and Systems, School of Computer Science and Engineering, Beihang University, and also an Assistant Professor with Peng Cheng Laboratory, Shenzhen, China. His research interests include Computer Graphics, Virtual Reality, Computer Vision and Robotics.
\end{IEEEbiography}






\end{document}